\documentclass[letterpaper, 10 pt, conference]{ieeeconf}  
\usepackage[hidelinks]{hyperref}
\usepackage{algorithm, algorithmic}
\usepackage{amsmath}
\usepackage[lmargin=2cm, rmargin=2cm, top=1.5cm, bottom=2cm]{geometry}

\usepackage{caption}
\usepackage{booktabs}
\usepackage{makecell}
\usepackage{siunitx}
\usepackage{tcolorbox}
\usepackage{pdfpages}
\usepackage{graphicx}
\usepackage{subcaption}
\usepackage{xspace}
\newcommand{\bfsection}[1]{\noindent\textbf{#1}}

\newcommand{\tabref}[1]{Table~\ref{#1}}
\newcommand{\figref}[1]{Fig.~\ref{#1}}
\newcommand{\eqnref}[1]{\text{Eq.}~(\ref{#1})}
\newcommand{\secref}[1]{\S\ref{#1}}
\newcommand{\appref}[1]{Appendix~\ref{#1}}

\newcommand{\algname}{\textsc{RELAX}\xspace}

\IEEEoverridecommandlockouts                              

\title{\LARGE \bf
RELAX: Reinforcement Learning Enabled 2D-LiDAR Autonomous System for Parsimonious UAVs
}

\author{Guanlin Wu$^{1}$, Zhuokai Zhao$^{2}$, and Yutao He$^{3}$
\thanks{$^{1}$Guanlin Wu is with the School of Data Science,
        The Chinese University of Hong Kong, Shenzhen, 
        Shenzhen, GD 518172, China
        {\tt\small guanlinwu@link.cuhk.edu.cn}}%
\thanks{$^{2}$Zhuokai Zhao is with the Department of Computer Science,
        University of Chicago, 
        Chicago, IL 60637, USA
        {\tt\small zhuokai@uchicago.edu}}%
\thanks{$^{4}$Yutao He is with the Computer Science Department,
        University of California, Los Angeles, 
        Los Angeles, CA 90095, USA
        {\tt\small yutao@cs.ucla.edu}}%
}

\begin{document}

\maketitle

\begin{abstract}

Unmanned Aerial Vehicles (UAVs) have become increasingly prominence in recent years,
finding applications in surveillance, package delivery, among many others. 
%
Despite considerable efforts in developing algorithms that enable UAVs to
navigate through complex unknown environments autonomously, they often require
expensive hardware and sensors, such as RGB-D cameras and 3D-LiDAR, leading to a persistent trade-off between performance 
and cost.
To this end, we propose \algname, a novel end-to-end autonomous framework that is
exceptionally cost-efficient, requiring only a single 2D-LiDAR 
to enable UAVs operating in unknown environments.
Specifically, \algname comprises three components: a pre-processing 
\textit{map constructor}; an offline \textit{mission planner}; and a reinforcement 
learning (RL)-based \textit{online re-planner}.
Experiments demonstrate that \algname offers more robust dynamic navigation compared to 
existing algorithms, while only costing a fraction of the others.
%
%
The code will be made public upon acceptance.

\end{abstract}

\section{Introduction}
Unmanned Aerial Vehicles (UAVs), commonly known as drones, have 
gained immense importance and become a transformative technology 
across many application domains~\cite{rovira2022review}.
In addition to more commonly-known use cases such as military 
navigation~\cite{patil2020survey}, 
search-and-rescue~\cite{mishra2020drone}, and commercial package 
delivery~\cite{alvarado2021237}, UAVs are also widely used in 
metrology~\cite{wieczorowski2021use},
agriculture~\cite{ahirwar2019application}, and
mining~\cite{shahmoradi2020comprehensive} thanks to their compact sizes and 
relatively high cost-efficiency, especially when compared to the piloted aircraft.

Despite tasks from different applications pose different, often specific 
challenges, one of the key challenges shared across all domains is being able
to operate autonomously.
%
%
Specifically, it covers many aspects of the UAV operations, including environment 
perception, path planning and real-time dynamic obstacle avoidance.
It is especially important when UAVs are to operate under unknown or dynamically 
changing environments, where human control is unavailable.

UAV autonomous navigation algorithms require on-board sensors to understand the 
surrounding environments, and optimally navigates UAV to travel from one place to 
another~\cite{barnhart2021introduction}. 
Optimal navigation can be defined in terms of the length of the traveled
path~\cite{noreen2016optimal}, traveled time~\cite{kularatne2016time} and 
trajectory smoothness~\cite{xu2021new} while being 
collision-free~\cite{shin2020performance}.
Numerous efforts have been devoted to advance this field~\cite{jones2023path}.
However, existing solutions often require UAVs to equip with expensive sensor setups,
including multiple RGB-D
cameras~\cite{cheng2021neural, xu2023real, machines10100931, kim2022autonomous} 
or 3D LiDAR~\cite{qin2019autonomous}.
While these sensors can help build real-time 3D maps for better environment
representations, they significantly increase the cost of the autonomous system, 
as most RGB-D cameras are priced at a few hundreds dollars in average~\cite{ulrich2020analysis}, 
and 3D LiDAR often costs well above a thousand US dollars~\cite{van2021solid}.
Consequently, such high cost has become a primary factor preventing UAVs from wider 
adoptions~\cite{aggarwal2020path}.
Therefore, new solutions enabling UAVs to perform \textit{successful and reliable autonomous 
navigation while using much simpler and cheaper sensor setups} are urgently needed.

Simpler sensor configuration, which changes the system perception of the surrounding
world, poses many new challenges for all system components including surrounding environment
construction, path planning, dynamic obstacle avoidance, and often calls for an entire new 
system design.
Specifically, it introduces practical challenges in surrounding 
detection~\cite{ocando2017autonomous, murcia20183d}
and RL training~\cite{ding2020challenges}.
To this end, we introduce \textbf{R}einforcement Learning \textbf{E}nabled 2D-\textbf{L}iDAR 
\textbf{A}utonomous \textbf{S}ystem (\textbf{\algname}), an end-to-end autonomous system 
presenting novel algorithms addressing these intricacies, so that parsimonious UAVs that 
carry only one 2D-LiDAR sensor can navigate autonomously in unknown environments.
Specifically, \algname comprises three components: 
a \textit{map constructor}, which generates occupancy maps using 2D-LiDAR data; 
a \textit{mission planner}, which creates obstacle-free paths using these maps; 
and an \textit{online re-planner}, which addresses the dynamic obstacle avoidance.

The main contribution of this paper is that we propose \algname, \textit{the first UAV 
autonomous navigation system that requires only a single 2D-LiDAR to support the entire
UAV autonomous navigation pipeline}, which includes the initial environment mapping, offline 
planning and online re-planning for dynamic obstacle avoidance. 
To address the unique challenges that come with the less feature-rich sensor inputs, 
we propose novel algorithms to enhance the capability and generalizability of our framework.
Experiments shows that \algname achieves comparable successful rates as more expensive UAVs
navigation systems, at only a fraction of the cost.
In addition, we advocate \algname as a \textit{successful proof-of-concept and a platform that 
boosts future research} by releasing a real-time training suite in ROS-Gazebo-PX4 simulator, 
which supports easy adaptation of \algname algorithms into training future newly designed RL 
algorithms.
%
In other words, the idea of modularization behind the design of \algname brings larger 
potential for further improvement of its performance.
%
%


\section{Related Work}\label{sec:related_work}
%
%
Existing end-to-end UAV autonomous navigation systems leverage sensor (e.g. RGB-D, 3D-LiDAR) 
inputs to perceive and understand surrounding environment, then conduct path planning and 
automatic dynamic obstacle avoidance~\cite{elmokadem2021towards}.
Besides differences in the algorithmic aspects, sensor configurations also fundamentally
affect the overall design of the system architecture, as well as the specific algorithms
within each component.
In this section, we briefly discuss different UAV navigation systems that equip with different 
sensor configurations.

\bfsection{Vision-based UAV navigation systems.}
Vision-based systems that employ RGB or RGB-D images to capture the environment are arguably 
the most prevalent configuration in autonomous UAVs~\cite{lu2018survey}.
More specifically, RGB images are taken by monocular cameras, while RGB-D images refer 
to 3D representations of the world world that is captured by either binocular cameras or 
monocular camera with additional depth sensor.
%
%

Numerous efforts have been devoted to vision-based UAV systems.
For example, Engel et al.~\cite{engel2014scale} developed a quadrator carrying a
monocular camera that is capable of visual navigation in unstructured environments.
Although being low in cost, the proposed system does not support obstacle 
avoidance, which is a major disadvantage for many modern tasks.
As a result, many works choose to use binocular 
cameras~\cite{mao2019development, jingjing2019research}.
However, such systems are very prune to weather changes and are hard to operate at 
night, greatly limiting their working scenarios.

Because of the aforementioned disadvantages of monocular and binocular configurations, RGB-D 
which involves both RGB and infrared depth cameras quickly attracts many attentions, 
resulting in various UAV applications~\cite{bachrach2012estimation, xu2023real}.
While being effective, the use of RGB-D cameras inevitably increases both cost and
on-board computational requirement, posing limitations and preventing designs of
simple, low-cost and light-weight UAVs for wider adoptions.

\bfsection{LiDAR-based UAV navigation systems.}
Thanks to its robust performance under various weather and lighting conditions, 
LiDAR has quickly become the mainstream sensor in many modern UAV 
navigation systems~\cite{jeong2018evaluation, qin2019autonomous}.
LiDAR sensors can be divided into two categories: single-line and multi-line, 
where single-line scans one plane of the obstacles to obtain a 2D map, while 
multi-line scans multiple surfaces to obtain a 3D point cloud of the environment.
Based on the output types, single- and multi-line LiDAR are also called 2D and 3D 
LiDAR.

Attracted by the richer environment representations that 3D LiDAR produces, most
existing UAV autonomous systems utilize 3D LiDAR as the sensor 
configurations~\cite{qin2019autonomous, aldao2022lidar, liang2023autonomous}.
However, despite existing work's favor into 3D LiDAR, the rich 3D environmental 
representations might not be all necessary to perform UAV path planning 
and robust obstacle avoidance, leaving room for better cost-effective designs.
In other words, configurations that utilize 2D LiDAR, where we call a 
\textit{parsimonious configuration}, may achieve a more balanced trade-off 
between performance and cost.
For example, Gabriel et al.~\cite{Gabriel2023adaptive} leverage 2D LiDAR and
propose an adaptive path-planning solution that combines Rapidly Exploring 
Random Trees (RRT) and deep RL for the autonomous trajectory generation of UAVs 
in agricultural environments.
However, it does not depend on UAV-scanned data at all stages but rather leverages 
a comprehensive Python environment for its operations, failing to equip the system 
with efficient obstacle avoidance capabilities. 
Contrary to this, \algname prioritizes enhancing obstacle avoidance by mainly 
utilizing LiDAR data. 
More specifically, we employ the ROS-Gazebo-PX4 simulator for developmental purposes, 
incorporating a variety of algorithms aimed at overcoming different obstacles and 
ensuring the training's applicability in a real-time simulation setting.
%

%

\section{Methodology}\label{sec:methodology}
\algname is designed specifically for parsimonious UAVs, which are drones that lack odometers, 
RGB-D cameras, 3D-LiDAR, or gimbals systems, and only equip simple sensors, such as 
2D-LiDAR and inertial measurement unit (IMU). 
More specifically, \algname utilizes RPLiDAR\footnote{More details at 
https://www.slamtec.ai/product/slamtec-rplidar-a1/.}, a low-cost 2D laser scanner 
that performs $360$-degree scan within a certain range to produce 2D point clouds 
of the surrounding.
\begin{figure}[t]
    \centering
    \includegraphics[width=\linewidth]{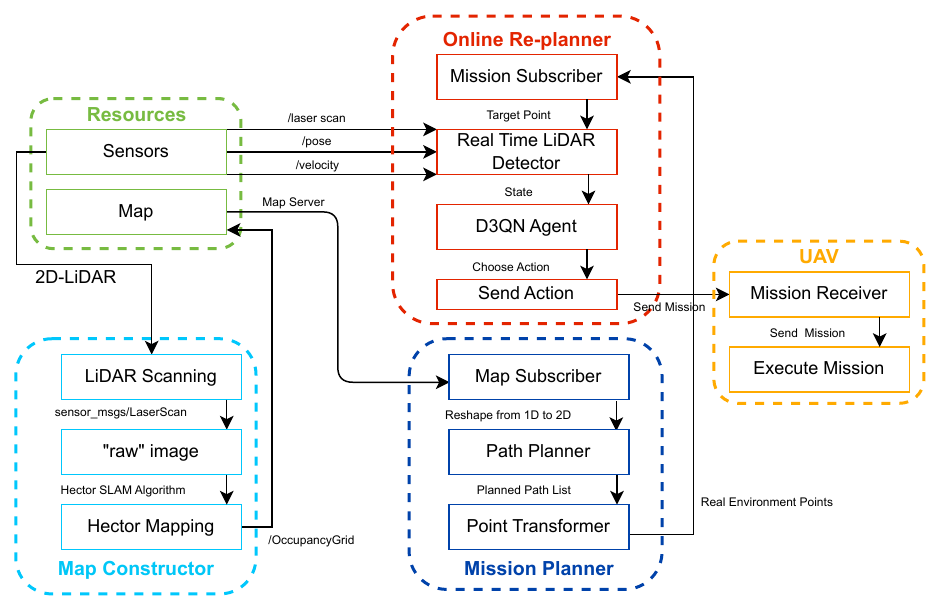}
    \captionsetup{font={small}}
    \caption{
    System overview: \algname starts from checking whether the occupancy grid map exists. 
    If there is no map, it will run \textit{map constructor} to enter the map constructing mode. 
    While we manually operate the drone to fly one complete circuit around the environment 
    at a specific altitude, \textit{map constructor} processes the data from 2D-LiDAR 
    and integrates these data to create an occupancy grid map. 
    This map is then sent back to \textit{resources} and available to other modules. 
    Next, \textit{mission planner} subscribes this map and use it to plan an obstacle-free 
    path from start to target and sends to \textit{online re-planner} for dynamic obstacle
    avoidance using real-time 2D-LiDAR inputs. 
    %
    %
    %
    %
    }
    \label{fig:gen_struc}
    \vspace{-0.1in}
\end{figure}

\algname consists of five modules, as shown in \figref{fig:gen_struc}. 
%
\textit{Resources} module contains necessary sensor outputs including point clouds 
captured by 2D-LiDAR, velocity and pose of UAV obtained from IMU, and the map generated 
by a \textit{map constructor}. 
Specifically, \textit{map constructor} synthesizes an occupancy map of the environment 
using point clouds from 2D-LiDAR.
\textit{Mission planner} provides an obstacle-free path from the 
starting point to the target position based on the occupancy map.
And \textit{online re-planner} navigates the drone (illustrated as the 
\textit{UAV} module) to move along this planned path and perform online 
re-planning to avoid dynamic obstacles. 
The ``dynamic obstacles" in this paper refers to the static obstacles that are not included in the map produced by \textit{map constructor}. 
%
%

Following~\cite{Gabriel2023adaptive}, we separate static path planning from online path
re-planning, with the underlying intuition that the environment does not undergo significant 
changes in a short period.
And separation in the different path planning stages significantly reduces the time cost.
%
%
We illustrate \textit{map constructor}, \textit{mission planner} and \textit{online re-planner} 
with details in \secref{subsec:map_constructor}, \secref{subsec:mission_planner},
and \secref{subsec:obstacle_handler}, respectively.

\subsection{Map Constructor}\label{subsec:map_constructor}
Map constructor leverages 2D-LiDAR in tandem with Hector-SLAM~\cite{kohlbrecher2011aflexible} 
to construct a grid occupancy map of the environment. 
%
%
%

\begin{figure}[t]
    \begin{subfigure}[h]{0.478\linewidth}
        \centering
        \includegraphics[width=\textwidth]{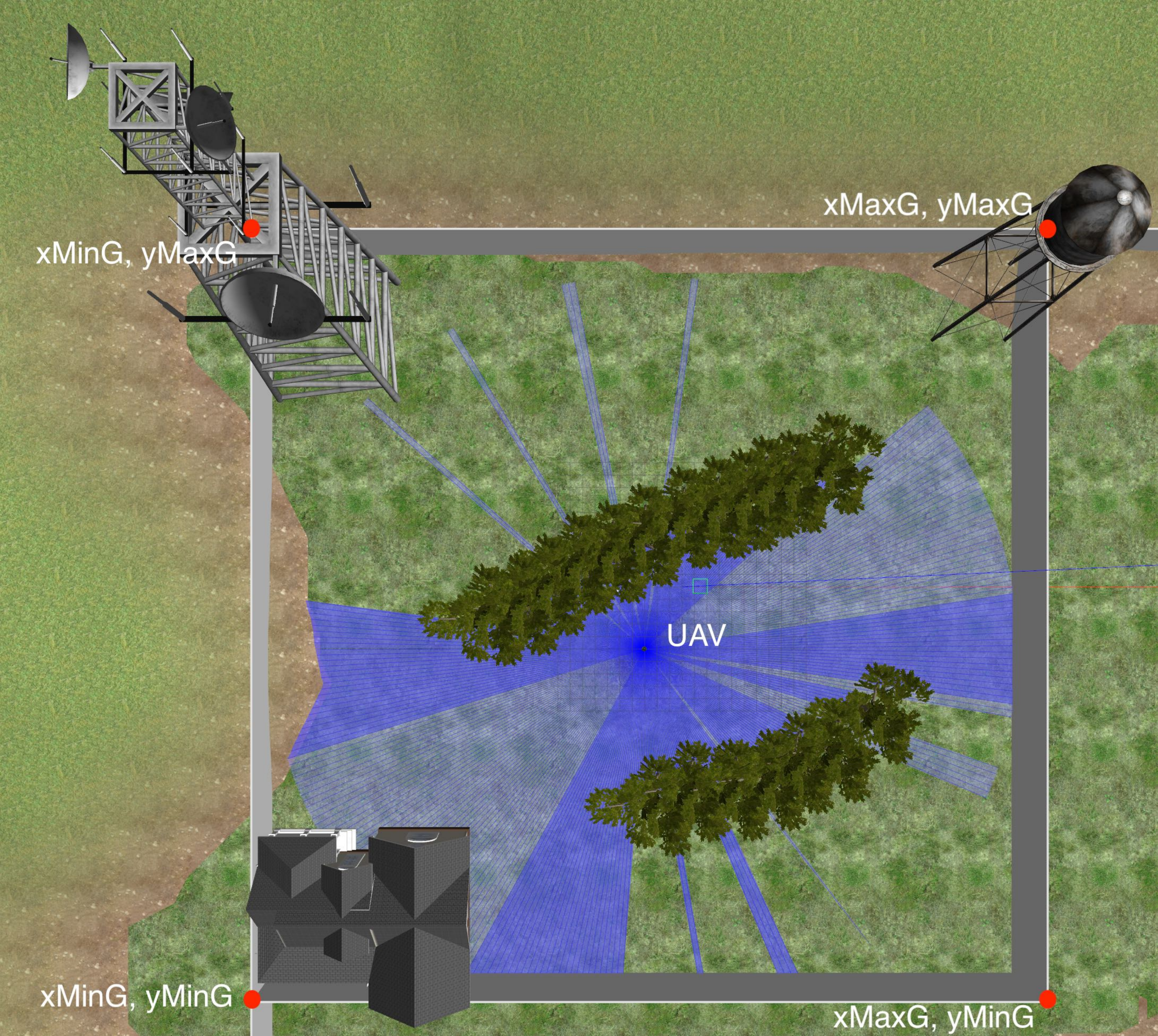}
    \end{subfigure}
    \begin{subfigure}[h]{0.50\linewidth}
        \centering
        \includegraphics[width=\textwidth]{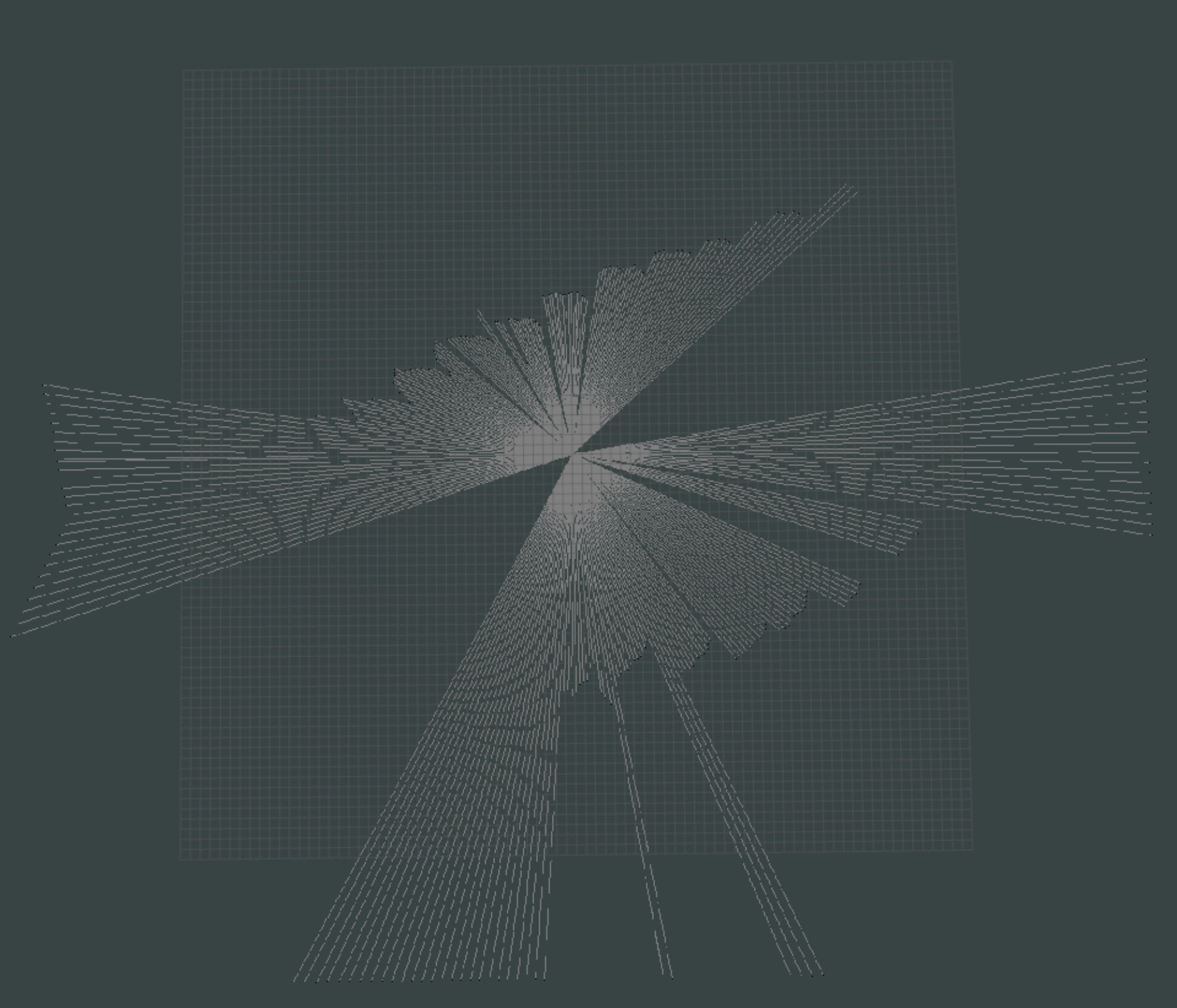}
    \end{subfigure}
    \captionsetup{font={small}}
    \caption{Left: environment of UAV at a particular position; 
    Right: ``raw" 2D-LiDAR scanning image of the left environment.}
    \label{fig:env_raw_lidar}
    \vspace{-0.1in}
\end{figure}

\bfsection{LiDAR scanning.}\label{subsubsec:lidar_scanning}
LiDAR scanning module generates raw images of the surrounding 
environment at a particular UAV position, as shown in the left of 
\figref{fig:env_raw_lidar}. 
In 2D-LiDAR system, the scanned images adopts a structure that 
aligns with $x$- and $y$-axis after a reshape operation performed on 
the one-dimensional LiDAR data array. 
Then the image is processed into an occupancy grid map, as shown in the right of
\figref{fig:env_raw_lidar}, where the level of confidence regarding obstacle 
existence is represented through dark (low) to light (high). 
While flying through the environment, the drone constantly generates 
``raw" images, contributing to the ongoing construction of the environment.

\begin{figure}[htbp]
    \centering
    \begin{subfigure}[h]{0.45\linewidth}
        \centering
        \includegraphics[width=\textwidth]{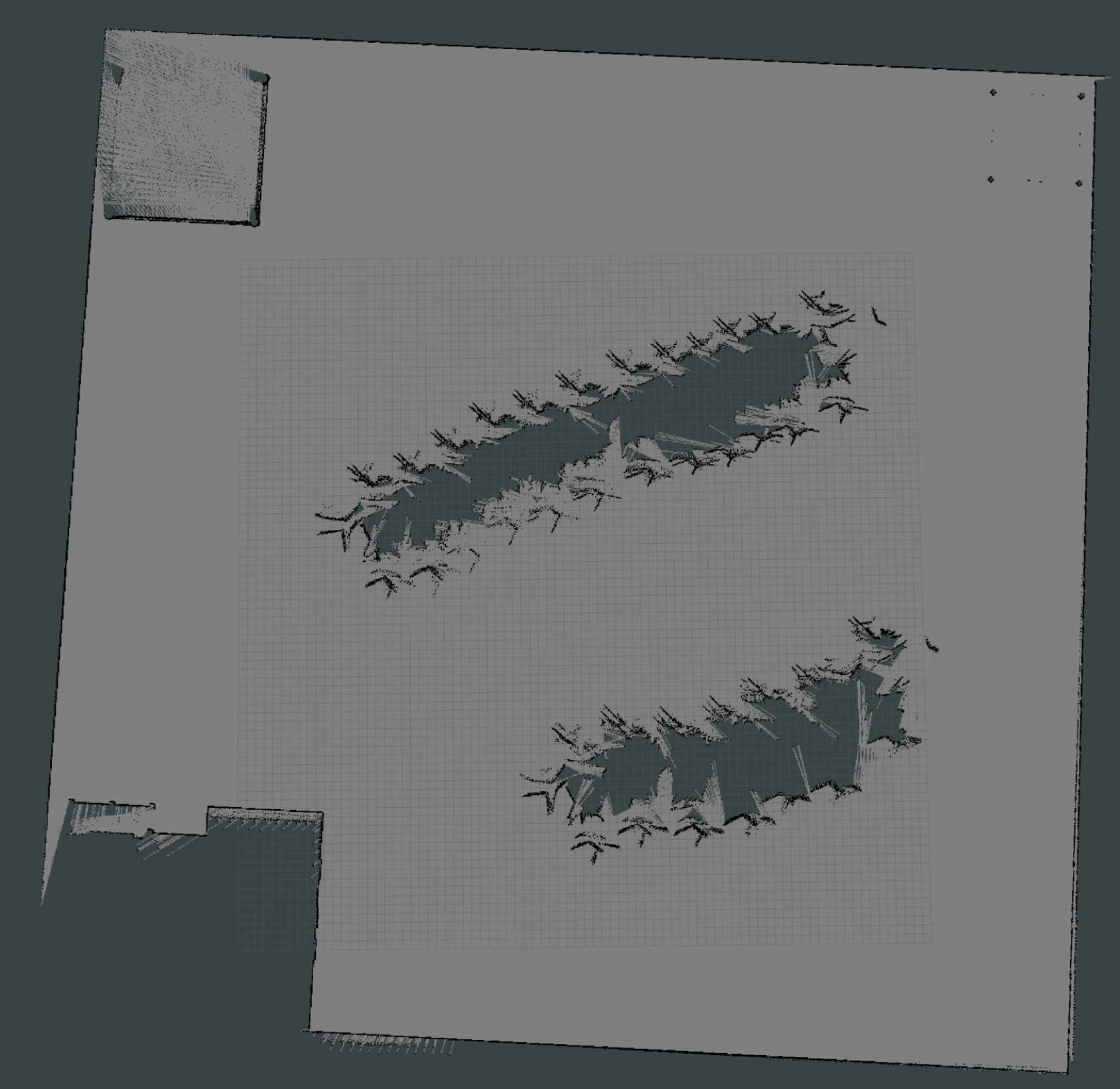}
    \end{subfigure}
    \begin{subfigure}[h]{0.465\linewidth}
        \centering
        \includegraphics[width=\textwidth]{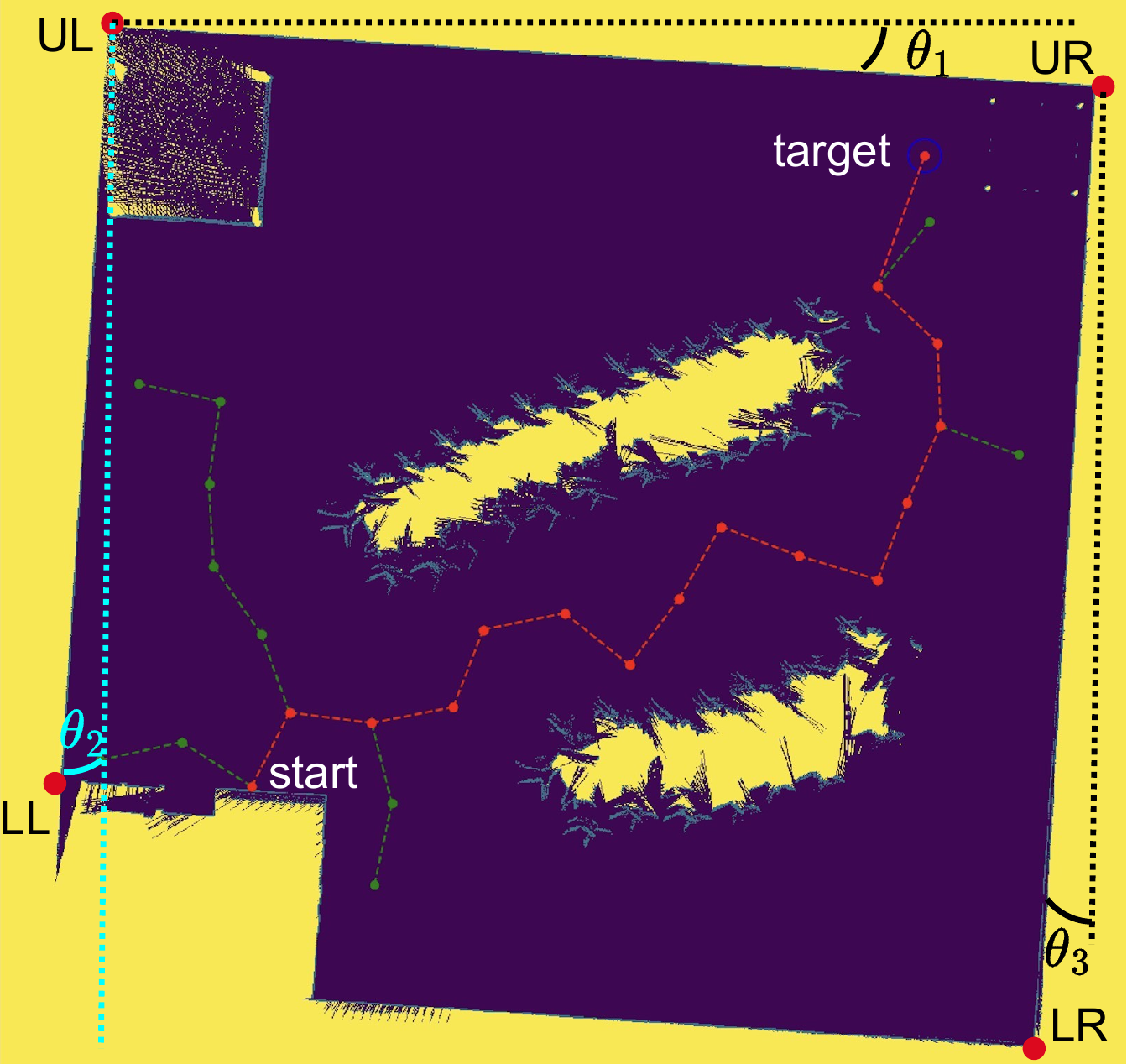}
    \end{subfigure}
    \captionsetup{font={small,stretch=1}}
    \caption{
    Left: occupancy map constructed by Hector-SLAM.
    Right: a path planning result based on the given occupancy map. 
    %
    }
    \label{fig:cons_map_plan_path}
    \vspace{-0.1in}
\end{figure}

\bfsection{Hector-SLAM.}\label{subsubsec:hector_slam}
%
%
%
Map constructor employs Hector-SLAM~\cite{kohlbrecher2011aflexible} to integrate 
all the LiDAR-scanned ``raw" images into a single map that represents the entire environment.
%
%
More specifically, Hector-SLAM operates across three primary phases, which are 
\textit{map access}, \textit{scan matching} and 
\textit{multi-resolution map representation}. 
In \textit{map access}, the initial occupancy grid map takes shape,
driving from the first ``raw" image. 
Next, \textit{scan matching} matches the ``raw" image taken at time $t$ to the 
previous occupancy grid map from $t-1$ through points correspondence.
%
%
To lower the risk of getting stuck in local optimal solution, Hector-SLAM 
applies \textit{multi-resolution map representation} to simultaneously keep 
different maps and update them based on pose estimations. 
%
%
The resulting map in our case is shown in the left of \figref{fig:cons_map_plan_path}.
%
%
\subsection{Mission Planner}\label{subsec:mission_planner}
Mission planner receives the occupancy map from map constructor, 
and plans an obstacle-free path from start to end. 
It includes two components, which are \textit{path planner} and 
\textit{point transformer}.
%
The complete algorithm of mission planner is detailed in \appref{alg:mission_planner}.

\bfsection{Path planner.}\label{subsubsec:path_planner}
Path planner is responsible for generating a collision-free path from start to 
end based on the static occupancy map.
In our case, this map refers to the output of map constructor.
Since the generation of such path does not depend on characteristics unique to 
parsimonious UAVs, any standard path planning algorithm should suffice.
For wider adaptability, lower execution time, and relatively optimal path, 
we use Rapidly Exploring Random Tree (RRT)~\cite{steven2001randomized}
in this paper to showcase the feasibility of our proposed framework.
An example path is shown in the right of \figref{fig:cons_map_plan_path}.
%
%

\bfsection{Point transformer.}\label{subsubsec:point_transformer}
To navigate UAV through real-life environment, a transformation is needed to 
convert the path from path planner into real-life coordinates.
%
%
To begin with, we initiate a rotation of the map, as shown in the right of 
\figref{fig:cons_map_plan_path}. 
The rotation angle emerges from the cumulative summation of three lines' shifting 
angles ($\theta_1$, $\theta_2$, $\theta_3$ in \figref{fig:cons_map_plan_path} 
right), where each bears a weight that minimizes potential errors.
After rotation, every intermediate point along the trajectory undergoes 
calculation based on the ratio between distances in occupancy grid map and 
their counterparts in real environment. 
Four example points, \texttt{UL}, \texttt{LL}, \texttt{UR}, 
and \texttt{LR}
are illustrated in
\figref{fig:cons_map_plan_path}, where their corresponding points in real-life 
environment are the \texttt{xMinG}, \texttt{xMaxG}, \texttt{yMinG} and \texttt{yMaxG} 
in \figref{fig:env_raw_lidar}.
%

%
Let $\theta$ denotes the weighted sum of individual-axis rotation angles, 
$r$ denote the distance between origin and point $p$, and ($x_{pr}, y_{pr}$) 
be the $x$ and $y$ of $p$ after rotation, we have the real-life environment 
coordinates ($x^{\text{new}}_{p}, y^{\text{new}}_{p}$):
\begin{small}
\begin{align}
    \text{x}^{\text{new}}_{p} &= \left(r \cdot cos(\theta) + x_{pr}\right) 
    \cdot \frac{\texttt{xMaxG} - \texttt{xMinG}}{\sqrt{(x_{ur} - x_{ul})^2 + (y_{ur} - y_{ul})^2}} \label{eq:xnew} \\
    \text{y}^{\text{new}}_{p} &= \left(r \cdot sin(\theta) + y_{pr}\right) 
    \cdot \frac{\texttt{yMaxG} - \texttt{yMinG}}{\sqrt{(x_{ur} - x_{lr})^2 + (y_{ur} - y_{lr})^2}} \label{eq:ynew}.
\end{align}
\end{small}

\subsection{Online Re-planner}\label{subsec:obstacle_handler}
As the multitude of dynamic obstacle scenarios makes it impractical to establish 
comprehensive avoidance rules, a learning-based planning algorithm is designed to 
perform autonomous obstacle avoidance in dynamic, unknown environment.
%
%
More specifically, we propose a novel RL-based online re-planner combining 
Double Deep Q-networks (DDQN)~\cite{van2016deep} and dueling 
architecture~\cite{wang2016dueling}.

\bfsection{Network structure.}\label{subsubsec:network_structure}
Dueling Double Deep Q-networks (D3QN) improved upon Deep Q-networks 
(DQN)~\cite{mnih2013playing} and Double Deep Q-networks (DDQN)~\cite{van2016deep}
by incorporating the dueling architecture~\cite{wang2016dueling}.
More specifically, it splits the Q-values estimations into two separate functions, 
namely a value function, $V(s)$, estimating the reward collected from state $s$; 
and an advantage function, $A(s, a)$, estimating if action $a$ is better than other 
actions at state $s$.
Both value and advantage functions are constructed with a set of dense
layers and are later combined to output Q-values for each action, with the 
combination operator shown in \eqnref{eq:d3qn}.
\begin{equation}\label{eq:d3qn}
    Q(s, a) = V(s) + \left(A(s, a) - \frac{1}{|A|}\sum_{a'}{A(s, a')}\right)
\end{equation}

\begin{figure}[htbp]
    \centering
    \begin{subfigure}[h]{0.475\linewidth}
        \includegraphics[width=\textwidth]{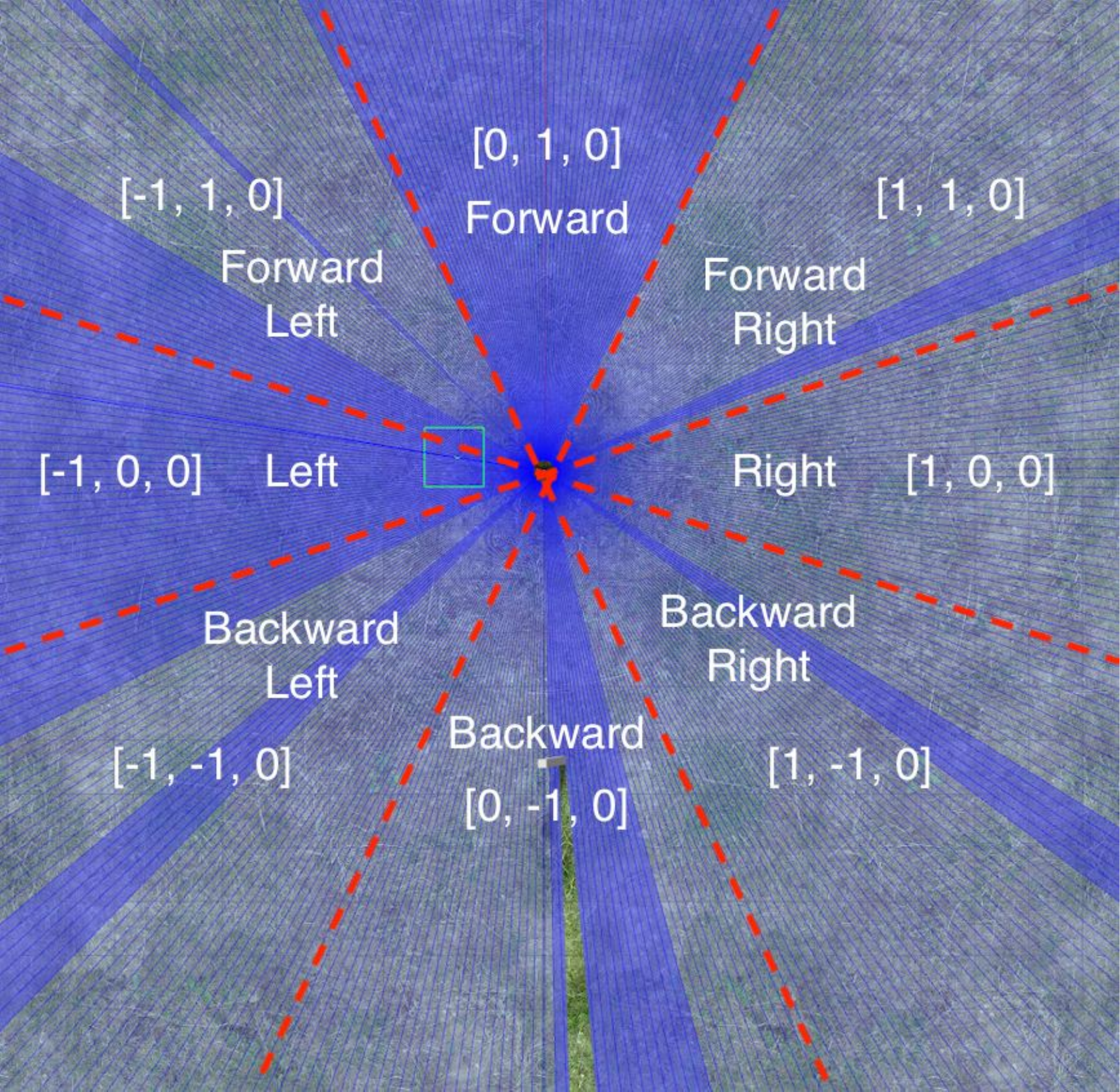}
    \end{subfigure}
    \begin{subfigure}[h]{0.45\linewidth}
        \includegraphics[width=\textwidth]{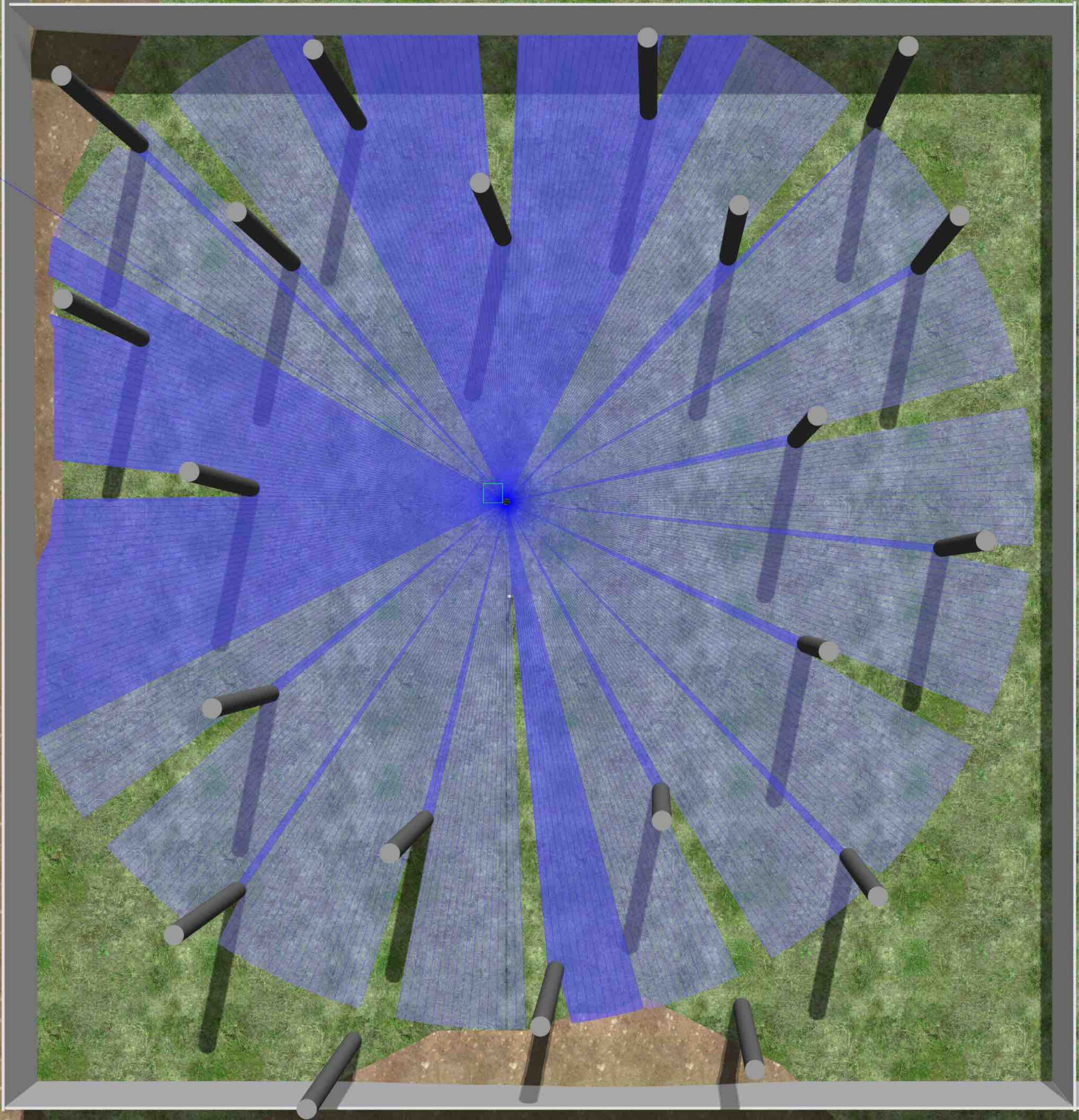}
    \end{subfigure}
    \captionsetup{font={small,stretch=1}}
    \caption{
    Left: state-action correspondence, where agent can choose or exclude action [$1, -1, 0$] 
    based on the distance.
    Right: training environment of the model.
    }
    \label{fig:sta_act_train_env}
    \vspace{-0.1in}
\end{figure}

\bfsection{State design.}\label{subsubsec:state_design}
%
%
We integrate the orientation vector spanning from the current location to the 
target, along with real-time LiDAR data, into our state design.
Specifically, real-time LiDAR data, which are 360-vectors representing
each degree, is partitioned into 8 sectors through thresholded min-pooling, where 
each corresponds to a specific direction, such as ``forward-left" or 
``forward-right", as shown in the left of \figref{fig:sta_act_train_env}.
More precisely, we have: 
\begin{equation}\label{eq:dist_min}
    d_i = \min\left(all\_dist \in region_i, det\_range\right), i \in [0,7]
\end{equation}
where $det\_range$ denotes the threshold value and $all\_dist$ represents the distances 
of all points (in our case is $\frac{360}{8} = 45$) in $region_i$. 
%
%
%
%
%
Let ($x_c, y_c, z_c$) and ($x_t, y_t, z_t$) denote the current and target position, 
we have the direction vector defined as:
\begin{equation}
    \left(x_d, y_d, z_d\right) = \left(x_t, y_t, z_t\right) - \left(x_c, y_c, z_c\right)
\end{equation}
Finally, the current state is defined as:
\begin{equation}\label{eq:state}
    state = [x_d, y_d, z_d, dist_0, dist_1, \dots, dist_6, dist_7]
\end{equation}
%

%
One of the biggest challenges using 2D-LiDAR is that the data may be extremely noisy
due to the disturbances from UAV maneuvers.
To address this challenge, we propose a novel data filtering mechanism, as
illustrated in Algorithm~\ref{algo:lidar_detection} in \appref{sec:appendix}, 
to enhance the accuracy of the acquired data. 
The core idea for judging whether this data is noisy is that within all potential 
actions, the greatest conceivable variation in distance between two states should be 
$\leq \sqrt{2} < 1.5$. 
The rule of $\sqrt{2}$ comes from our definition of action, which will be illustrated 
with more details later in this section.
%
Let $x_i$, $y_i$ denote the coordinate difference between step $i-1$ and step $i$ in $x$ and
$y$-axis, respectively, we have $\sqrt{\max|x_i| + \max|y_i|} \leq \sqrt{2}$.
On the other hand, if the disparity is larger than 1.5, it is deemed to be spurious noisy 
and is more carefully handled as shown in Algorithm~\ref{algo:lidar_detection}. 
The heuristic behind Algorithm~\ref{algo:lidar_detection} is that there is a very small 
likelihood of continuously obtaining noisy data more than $det\_range/2$ number of times 
within the same region. 
Thus, we maintain a $index\_list$ to record the number of times that noisy data occurred 
for each region and will dynamically decrease for getting data without noisy.
In addition, after getting noisy data, we will set the distance to $(det\_range/2)+2$ or subtract 1 from it depending on the corresponding number at $index\_list$.


\bfsection{Action space.}\label{subsubsec:action_space}
As we constrain UAV from moving vertically due to sensor limitations 
(RPLiDAR can only scan horizontally), the action space $A$, as illustrated in 
\eqnref{eq:action} and \figref{fig:sta_act_train_env} left, 
only contains 8 actions.
Specifically, we have
%
\begin{small}
\begin{equation}\label{eq:action}
    \begin{split}
        A = \{&[1, -1, 0], [1, 0, 0], [1, 1, 0], [0, 1, 0], \\
        &\quad [-1, 1, 0], [-1, 0, 0], [-1, -1, 0], [0, -1, 0]\} 
    \end{split}
\end{equation}
\end{small}
%

\bfsection{Reward function design.}
%
%
%
%
Temporal reward $r_t$ that trains the model to learn optimal actions reaching the
target point is illustrated in \eqnref{eq:reward}.
More specifically, let $d_{current}$ denote the distance between current and target 
position, and $d_{last}$ denote the distance between the previous and target 
position, we have:
%
\begin{small}
\begin{equation}\label{eq:reward}
	r_t =
    \begin{cases}
		3000, &d_{current} \leq 3 \\
		-3000, &num\_steps\_taken \geq max\_num\_steps  \\
        -50, &d_{current} > d_{last}    \\
        -4000, &collision
	\end{cases}
\end{equation}
\end{small}
%
%
And the final reward $r$ for each chosen action is simply:
%
\begin{equation}\label{eq:final_reward}
    r = r_t - \frac{d^2_{current}}{100}
\end{equation}
%
The underlying rationale for the configuration of the reward mechanism is to prioritize 
the drone's learning process in avoiding collisions as the paramount task, while also 
discouraging the behavior of continuously circling a point proximate to the target.

\bfsection{Check done.}
Ensuring timely notifications about the completion of an episode holds immense 
significance in RL training.
Since we incorporate real-time LiDAR data into state representation, the training 
must proceed in Gazebo-ROS-PX4 simulator, which introduces complexity to the reset 
process after collision events.
Upon drone's collision with an obstacle, automatic disarming occurs, and manual 
restarts of ROS, Gazebo, and PX4 are required for re-arming, rendering continuous 
training infeasible.
To expedite model convergence and streamline the intricate reset procedure, we 
devise the check done function as shown in \eqnref{eq:check_done}, which 
defines the completion of an episode when the distance between the UAV and the 
obstacle is less than a predefined collision threshold $col\_threshold$.
\begin{small}
\begin{equation}\label{eq:check_done}
	done =
	\begin{cases}
		True, &d_{current} \leq 3 \\
		True, &\lvert x_c \rvert > \lvert limit\_x\rvert \, or \, \lvert y_c \rvert > \lvert limit\_y\rvert \\
        True, &counter \geq step\_threshold \\
        True, &\exists i \in [0,7] \, s.t. \, dist_i \leq col\_threshold \\
        False, &else
	\end{cases}
\end{equation}
\end{small}

\bfsection{Reset.}\label{subsubsec:reset}
Reset operation is the guarantee of a well-trained model. 
%
%
During training, at the end of each episode, the drone will be set to 
($0, 0, 4.4$) in Gazebo world. 
However, according to the settings of ROS, the position of drone will not 
immediately be set to ($0, 0, 4.4$). 
It will either be directly set to ($0, 0, 4.4$) after some time, which depends on 
the distance between the previous position of drone and ($0, 0, 4.4$), or 
decreasing from the previous position to ($0, 0, 4.4$) step by step. 
For example, if the drone was at position ($5, 6, 4.4$), after setting to 
($0, 0, 4.4$) in Gazebo world, the position of drone in ROS topic might change as 
$(5,6,4.4) \rightarrow (4,6,4.4) \rightarrow (3,6,4.4) \rightarrow ... \rightarrow (0,0,4.4)$.

This setting raises a fatal problem: before the position of drone in ROS becomes 
($0, 0, 4.4$), the drone will move uncontrollably and has high risk of colliding 
with obstacles during this period. 
To solve this problem, we propose a novel reset algorithm as shown in 
Algorithm~\ref{algo:reset} in \appref{sec:appendix}, in which ($x_c, y_c, z_c$) 
denotes the current drone position in ROS topic.
The core idea of the Algorithm~\ref{algo:reset} is to control the movement of drone 
into a specific range of space, among which we can guarantee no collision will happen.
The parameters such as a$_{thr}$, b$_{thr}$, offset$_a$, offset$_b$ are manually set 
to serve this purpose and should be modified when training environment is different.

\bfsection{Training details.}\label{subsubsec:training_details}
%
To ensure the model learns a policy that is independent of absolute positions 
(specified as $x, y, z$), the target position for each episode will be generated 
randomly within a predetermined range for $x$ and $y$, while maintaining $z$ at a 
constant value.
The complete training procedure, which summarizes the core algorithm of our proposed
\algname, is illustrated in Algorithm~\ref{algo:d3qn}.
Detailed hyperparameter values are shown in \tabref{tab:train_params} in \appref{sec:appendix}.
\setlength{\textfloatsep}{6pt}
\begin{algorithm}[htbp]
    \caption{RELAX}
    \label{algo:d3qn}
    \begin{algorithmic}[1]
    \renewcommand{\algorithmicrequire}{\textbf{Input:}}
    \renewcommand{\algorithmicensure}{\textbf{Output:}}
    \REQUIRE limits, start, max\_vel, max\_acc, max\_jerk, det\_range 
    \ENSURE  None
    \FOR {Number of Episodes}
        \STATE ($x_t, y_t, z_t$) $\gets$ randomly generated target position
        \STATE score $\gets 0$, counter $\gets 0$, Drone take off and go to start, last\_req $\gets Time.now()$
        \STATE Initial state $s_0$  $\gets$ LiDAR Data after \textcolor{blue}{Alg.\ref{algo:lidar_detection} LiDAR Data Filtering}  
        \WHILE {not done}
            \IF {drone.armed \AND $Time.now() -$ last\_req $>6$}
                \STATE Select an action $a_t$ with $\epsilon$-greedy algorithm
                \STATE Drone execute the action $a_t$, detect\_flag $\gets True$ \\
                \STATE New state $s_{t+1} \gets$ LiDAR Data after \textcolor{blue}{Alg.\ref{algo:lidar_detection} LiDAR Data Filtering}
                \STATE done $\gets$ \textcolor{blue}{Eq.\ref{eq:check_done}} Check Done, Reward $r_t \gets$ \textcolor{blue}{Eq.\ref{eq:final_reward}} Reward Function, score $+= r_t$
                \STATE Push ($s_t$, $a_t$, $r_t$, $s_{t+1}$, done) in memory buffer
                \IF {size(memory buffer) $\geq B$}
                    \STATE Sample $B$ transitions from memory buffer
                    \STATE Every $f_u$ steps update the $\theta_t$ with $\theta_p$, $\alpha_t$ with $\alpha_p$, $\beta_t$ with $\beta_p$
                    \STATE Do forward operation as \textcolor{blue}{Eq.\ref{eq:d3qn}} for $Q_{policy}$ and $Q_{target}$, get q\_pred, q\_next and q\_eval
                    \STATE q\_target = $r_t + \gamma*$q\_next
                    \STATE Update the parameter $\theta_p, \alpha_p, \beta_p$ for $Q_{policy}$ on loss (q\_target, q\_pred)
                \ENDIF
                \STATE $s_t = s_{t+1}$, counter $+= 1$
            \ELSE
                \STATE Maintain connection with drone
            \ENDIF
        \ENDWHILE
        \STATE Call \textcolor{blue}{Alg.\ref{algo:reset} Reset}
    \ENDFOR
    \end{algorithmic} 
\end{algorithm}
To emphasize the novelty of our proposed 2D-LiDAR UAV system \algname, we summarize the 
differences between the most updated existing frameworks, to our best knowledge, 
and ours in \tabref{tab:system_comparison}.
%
We see that \algname presents several advantages than the others.
Firstly, the mapping function, as we illustrated in \secref{subsec:map_constructor}, 
enables the use of our system in a wide range of environments without the need of 
manually reconstructing the environments.
Secondly, the ability to conduct real-time training in Gazebo-ROS-PX4 simulator ensures 
the learning of dynamic obstacle avoidance. 
More specifically, the dynamic obstacle avoidance utilizing real-time LiDAR data, as
illustrated in \secref{subsubsec:state_design}, improves significantly when compared to 
existing system~\cite{Gabriel2023adaptive} that utilizes only $(x, y, z)$ positions. 
%
\begin{table}[htbp]
    \sisetup{group-minimum-digits = 2}
    \centering
    \captionsetup{font={small,stretch=1}}
    \caption{Comparison between RELAX and other 2D-LiDAR UAV Frameworks: PP means path planning algorithm. Alg means the RL algorithm. 
    DOA means dynamic obstacle avoidance. TrInSim means real-time training in Gazebo-ROS-PX4 simulator and H-S means Hector-SLAM algorithm.}
    \label{tab:system_comparison}
    \begin{tabular}{ccccccS[table-format=5]} 
    \toprule
    \thead{Framework} & \thead{Mapping} & \thead{PP} & \thead{Alg} & \thead{DOA} & \thead{TrInSim} \\
    \midrule
    Gabriel's \cite{Gabriel2023adaptive} & - & RRT & DQN & - & -\\
    \textbf{RELAX (Ours)} & \textbf{H-S} & \textbf{RRT} & \textbf{D3QN} & $\surd$ & $\surd$ \\
    \bottomrule
    \end{tabular}
\end{table}

\section{Experiment -- A Case Study}\label{sec:experiment}
In this section, we demonstrate \algname on addressing a real-life challenge within the 
agricultural context, specifically catering to scenarios where farmers seek to do 
equipment checks during nocturnal hours or after extreme weathers such as storms. 
%
%
%

\bfsection{Hardware and software setup.}
The experiment is conducted on a desktop with Intel Core-i5-13400 CPU, Nvidia 
GeForce RTX4070Ti GPU and 64GB of RAM. 
The operating system is Ubuntu 20.04 bionic. 
The simulator is executed on Gazebo 11, ROS Noetic and PX4 v1.12.3.. 

\begin{figure}[t]
    \centering
    \begin{subfigure}[h]{0.43\linewidth}
        \includegraphics[width=\textwidth]{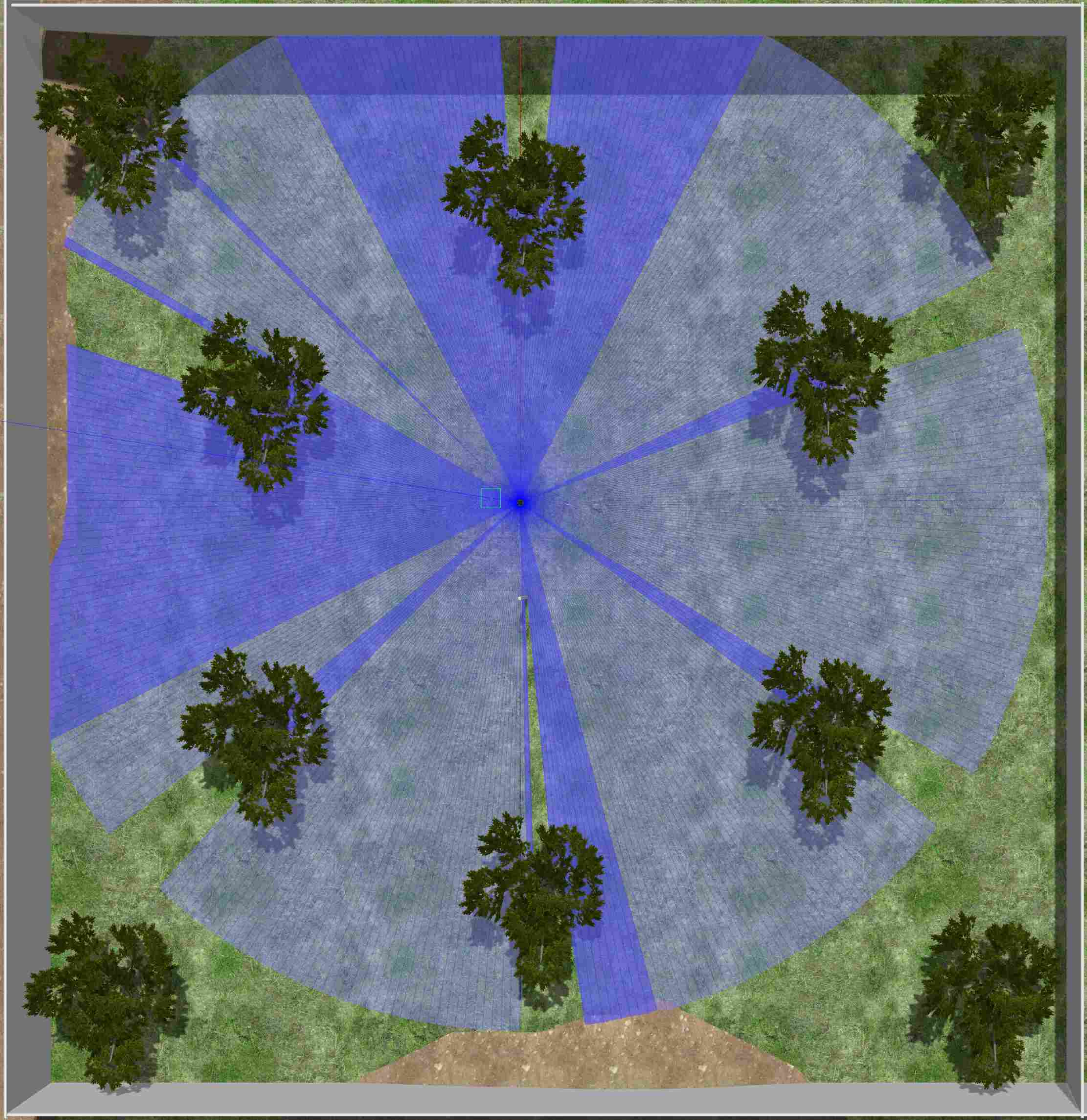}
    \end{subfigure}
    \begin{subfigure}[h]{0.55\linewidth}
        \includegraphics[width=\textwidth]{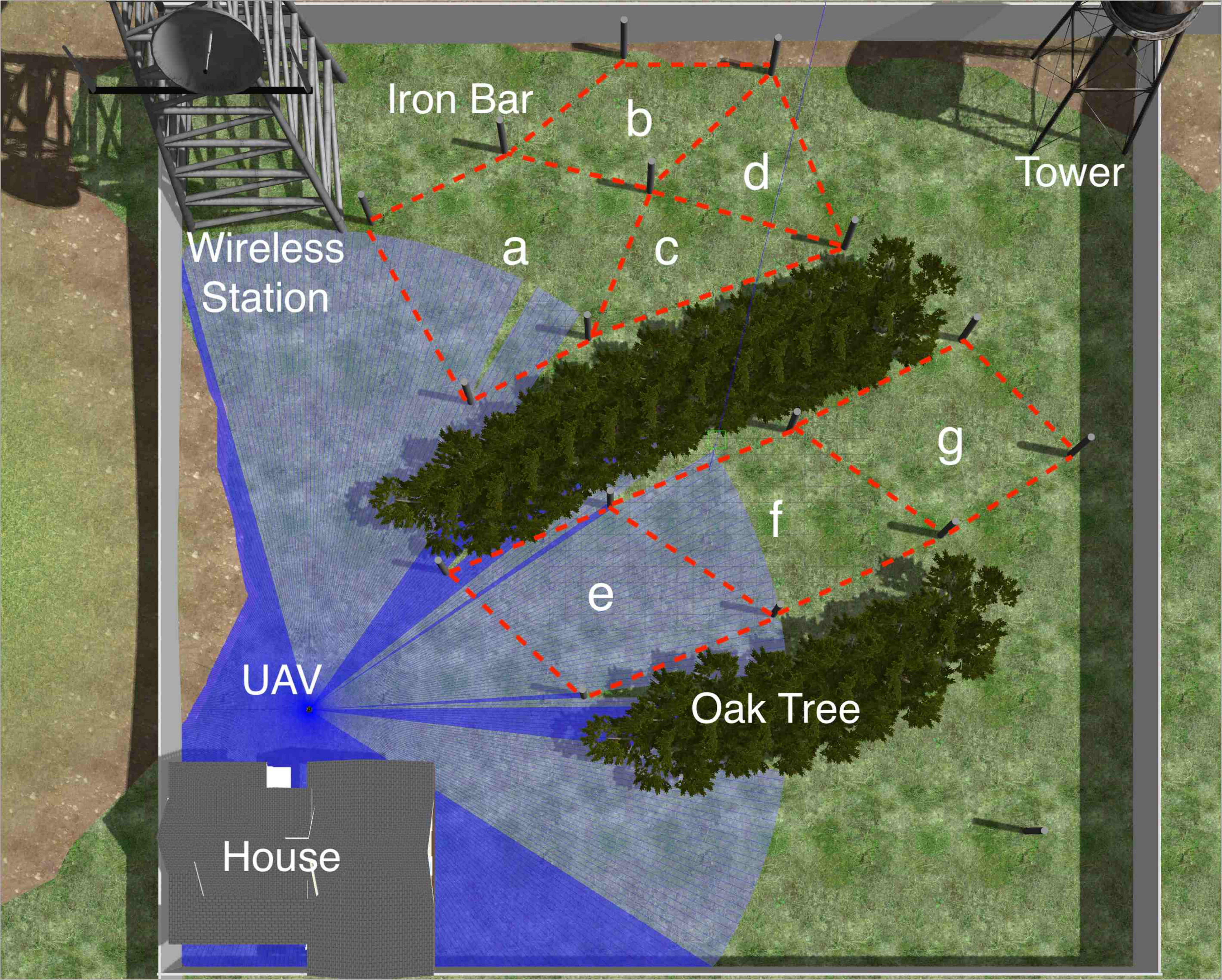}
    \end{subfigure}
    \captionsetup{font={small,stretch=1}}
    \caption{
    Left: training environment used for fine-tuning after training in environment shown 
    in \figref{fig:sta_act_train_env} right.
    Right: a typical farmland environment, where delineated regions are labeled as 
    $a, b, c$, and etc., and are separated by the movable iron bars 
    and wires (shown as red dotted lines). 
    The objective of the case study is to navigate a parsimonious UAV from 
    house to tower for nocturnal inspections.
    }
    \label{fig:finetune_env_test_env}
    \vspace{-0.1in}
\end{figure}

\bfsection{Experimental environment setup.}
We firstly train our RL model in environment shown in \figref{fig:sta_act_train_env} right and fine-tune it in environment shown in \figref{fig:finetune_env_test_env} left.
To examine the feasibility of our proposed framework, we established a test 
environment illustrated in \figref{fig:finetune_env_test_env} right. 
The agricultural land is divided into smaller zones by movable iron 
bars and wires to facilitate diverse crop cultivation. 
However, the dynamic nature of these iron bars, subject to seasonal rearrangements 
by farmers to accommodate varying crop types, presents noteworthy challenges, in 
which the static path planning based on a pre-scanned map becomes impractical.
The iron bars, which are not shown during the map construction stage, are considered as 
dynamic obstacles in our experiments.
%

%
%
%
%
%

\bfsection{Results.}
To comprehensively check the generalizability of our trained model, we conduct 50 run each 
with around 20 iron bars distributed randomly in the testing environment and report 
the average success rate.
The results are shown in \tabref{tab:results}, where the better variant of \algname (with D3QN) 
achieves an average success rate of 90\%, which is 8 times higher than the other 2D-LiDAR based 
algorithm, Gabriel's alg~\cite{Gabriel2023adaptive}, making \algname a practically usable 
solution in such agricultural applications. 
Additionally, the performance of \algname is on par with other state-of-the-art algorithms 
requiring 3D LiDAR and RGB-D cameras, such as Deep PANTHER~\cite{tordesillas2023deep} 
and FAST-LIO~\cite{kong2021avoiding}, showcasing \algname's competitiveness while keeping 
the total cost much lower. 
%
%
%
%
%
\begin{table}[t]
    \centering
    \captionsetup{font={small,stretch=1}}
    \caption{Performance Comparison between several algorithms and ours: For SPP (Static Path Planning), we mean planning a path on the map without iron bars, which mainly illustrates the performance difference between RRT and other algorithms. 
    For ORP (Online Re-Planning), it denotes the time needed for online re-planning to avoid the dynamic obstacles based on real-time 2D-LiDAR data, at which Genetic Algorithm \cite{Jinn2006tuning}, Dijkistra's Algorithm \cite{Dijkistra1959dijkistra}, and Gabriel's Algorithm \cite{Gabriel2023adaptive} are NOT able to handle. 
    Deep PANTHER \cite{tordesillas2023deep} is based on camera and FAST-LIO is based on 3D-LiDAR \cite{kong2021avoiding}. 
    Since the inference time of a trained RL model for each step differences a lot depending on state, we just record the range of time needed for one-time inference for illustration purpose. 
    For success rate, it illustrates the percentage of dynamic obstacles (iron bars) the drone avoided during the path it took from starting point to the target. 
    %
    %
    The average success rate of 50 tests for each algorithm are shown. 
    The bold number in each column illustrates achieving best performance for this criterion among all algorithms.}
    \label{tab:results}
    \begin{tabular}{ccccS[table-format=5]} 
    \toprule
    \thead{Algorithm} & \thead{Time(SPP)} & \thead{Time(ORP)} & \thead{Success Rate} \\
    \midrule
    Genetic Alg \cite{Jinn2006tuning} & 17.4s & - & 12\% \\
    Dijkstra's Alg \cite{Dijkistra1959dijkistra} & \textbf{4.8s} & - & 8\% \\
    Gabriel's Alg \cite{Gabriel2023adaptive} & 13.5s & - & 10\% \\
    Deep PANTHER \cite{tordesillas2023deep} & - & [0.01, 0.03] & \textbf{100\%} \\
    FAST-LIO \cite{kong2021avoiding} & - & \textbf{[0.003, 0.013]} & 98\% \\
    \textbf{RELAX (DQN)} & 13.5s & [0.0003, 0.1] & 82\%  \\
    \textbf{RELAX (D3QN)} & 13.5s & [0.0003, 0.05] & 90\% \\
    \bottomrule
    \end{tabular}
\end{table}

\begin{figure}[t]
    \centering
    \includegraphics[width=0.15\textwidth, angle=270]{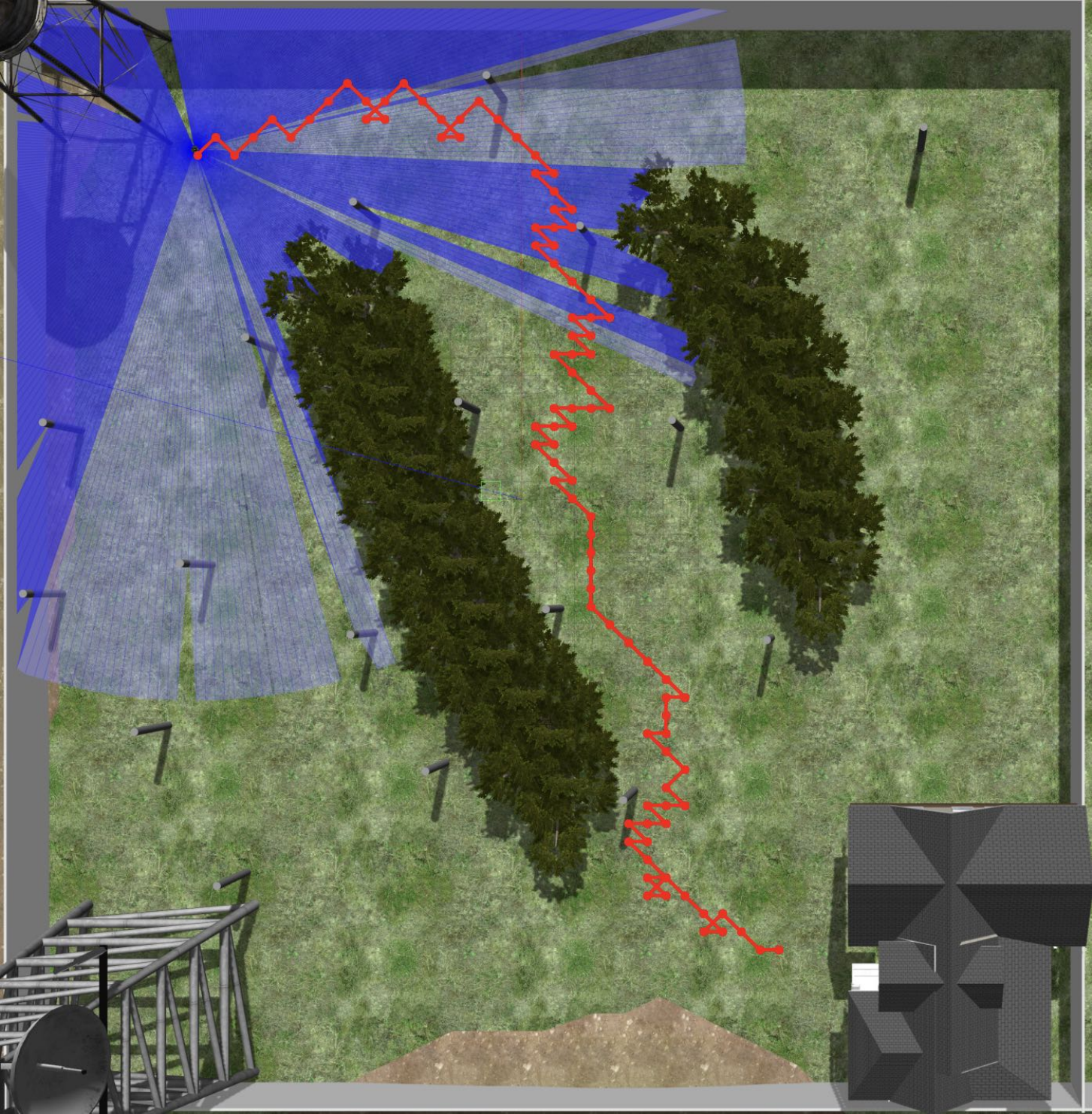}
    \includegraphics[width=0.15\textwidth, angle=270]{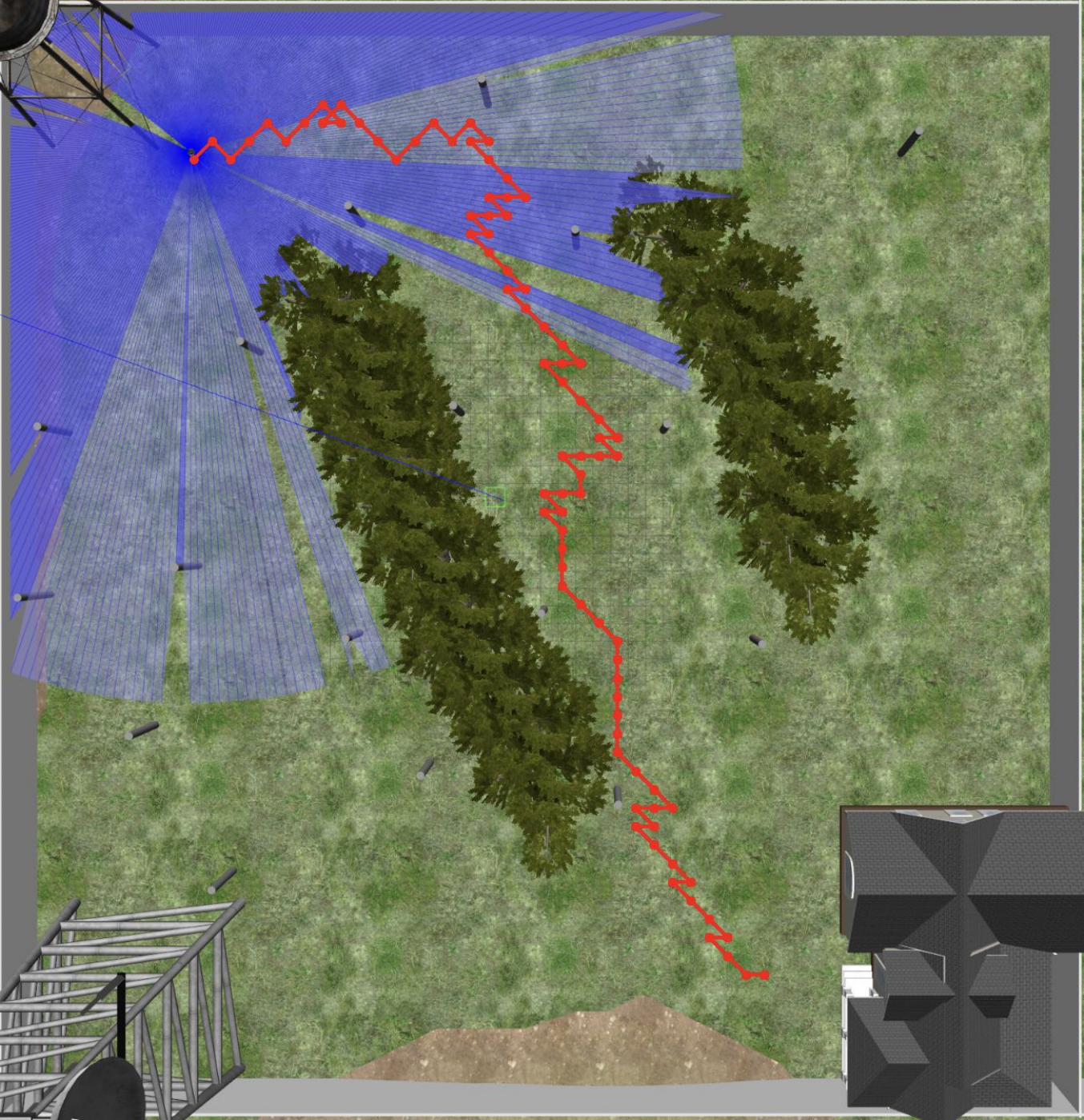}
    \includegraphics[width=0.15\textwidth, angle=270]{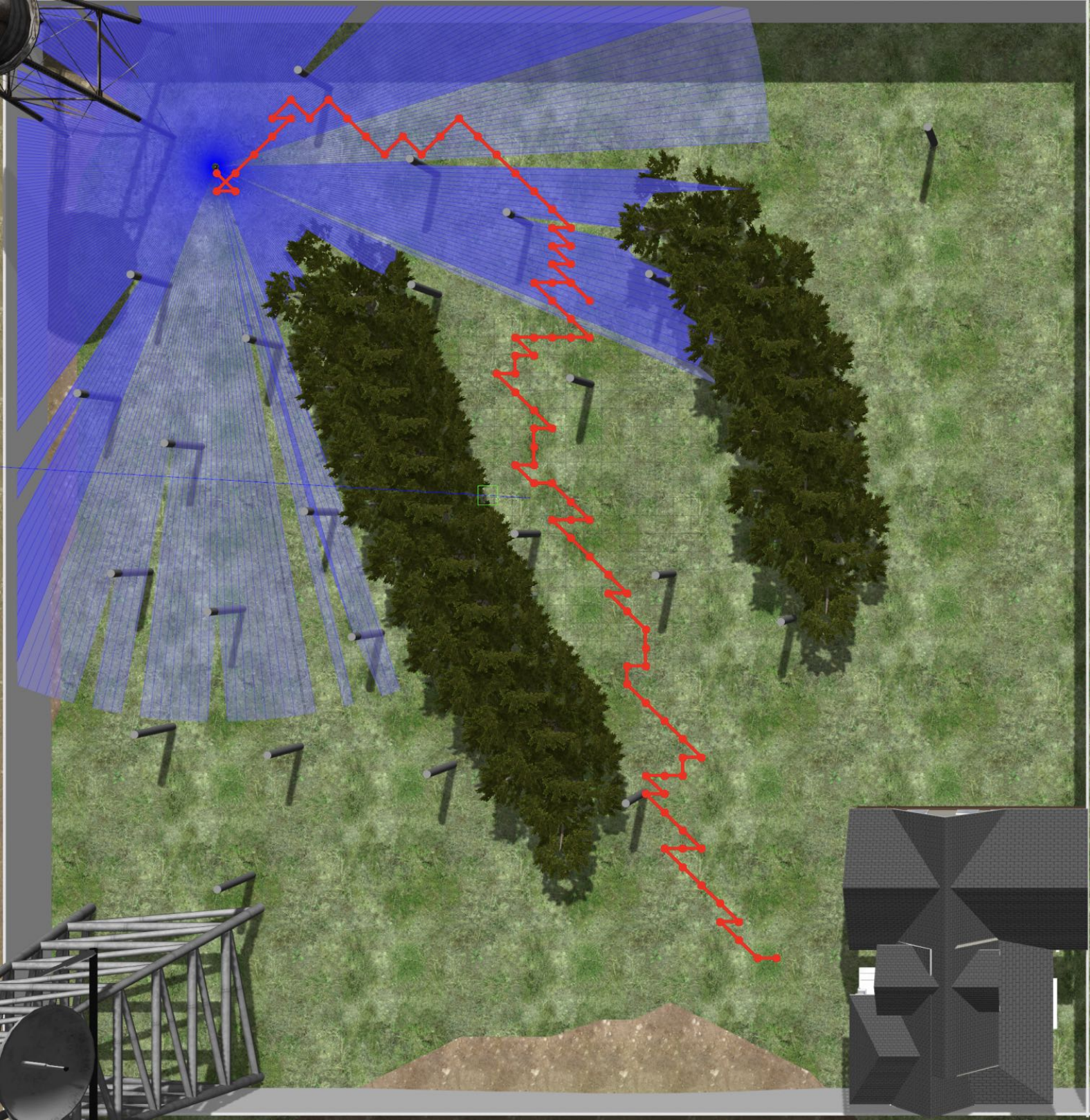}
    \includegraphics[width=0.15\textwidth, angle=270]{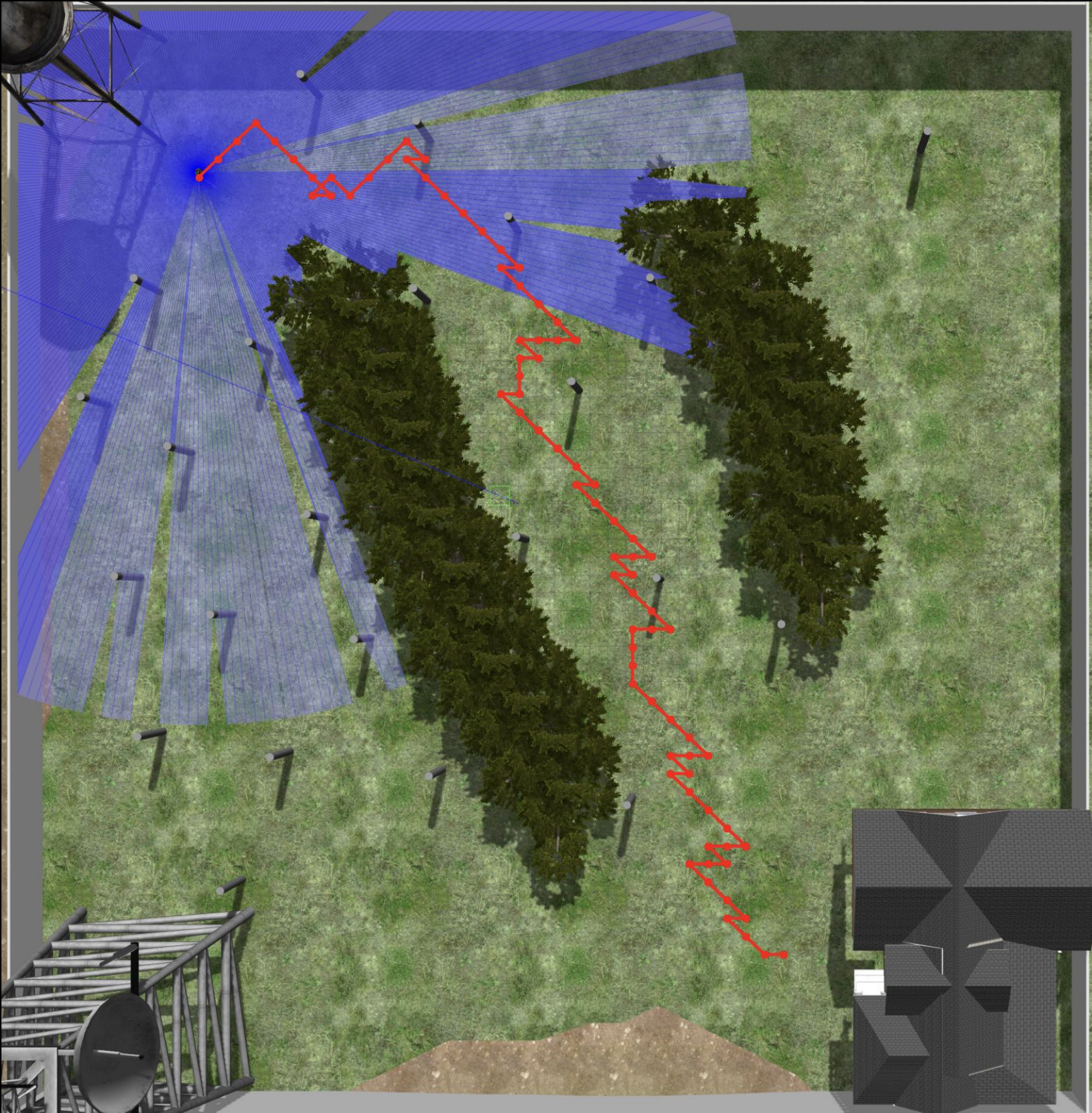}
    \includegraphics[width=0.15\textwidth, angle=270]{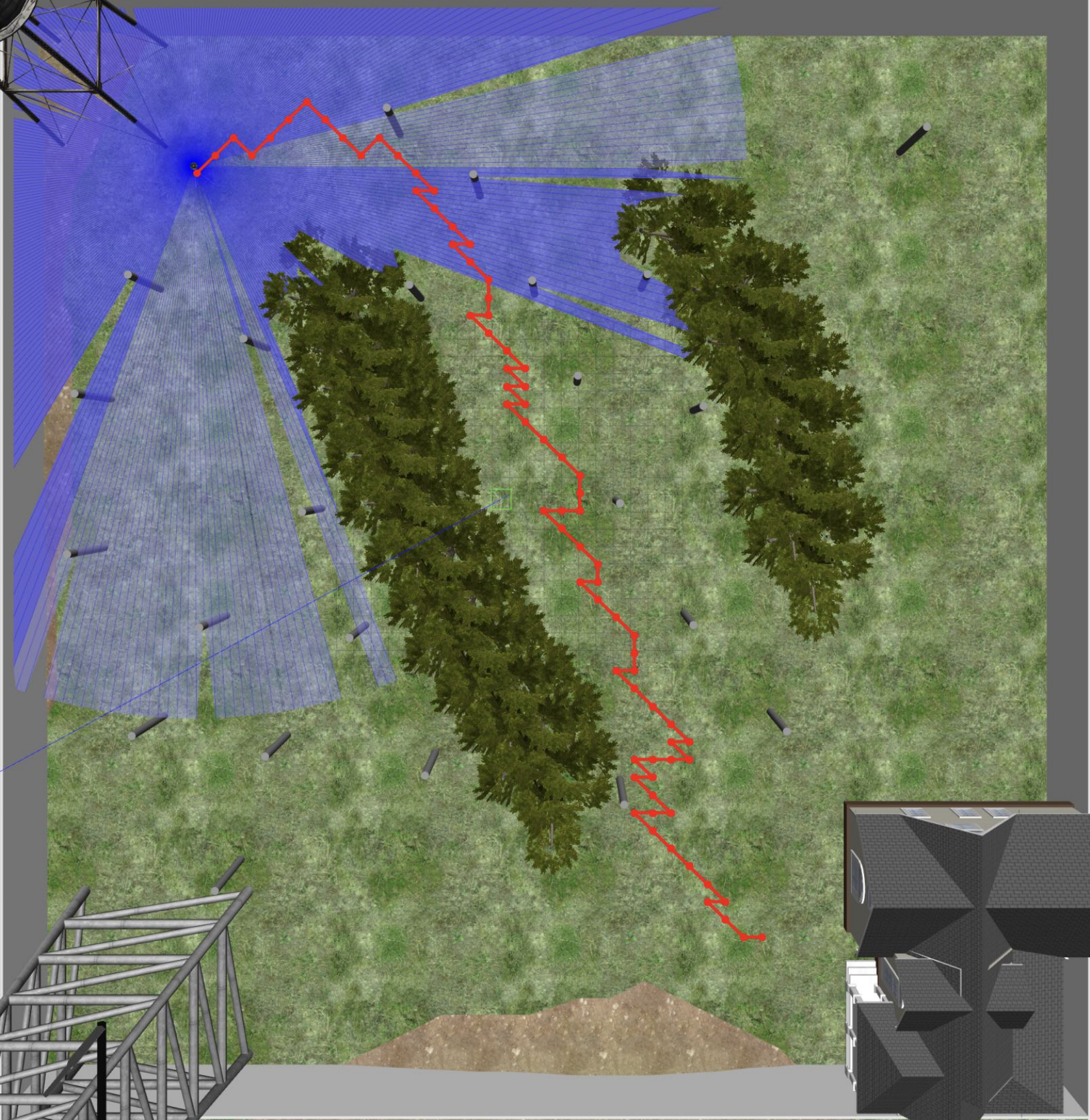}
    \includegraphics[width=0.15\textwidth, angle=270]{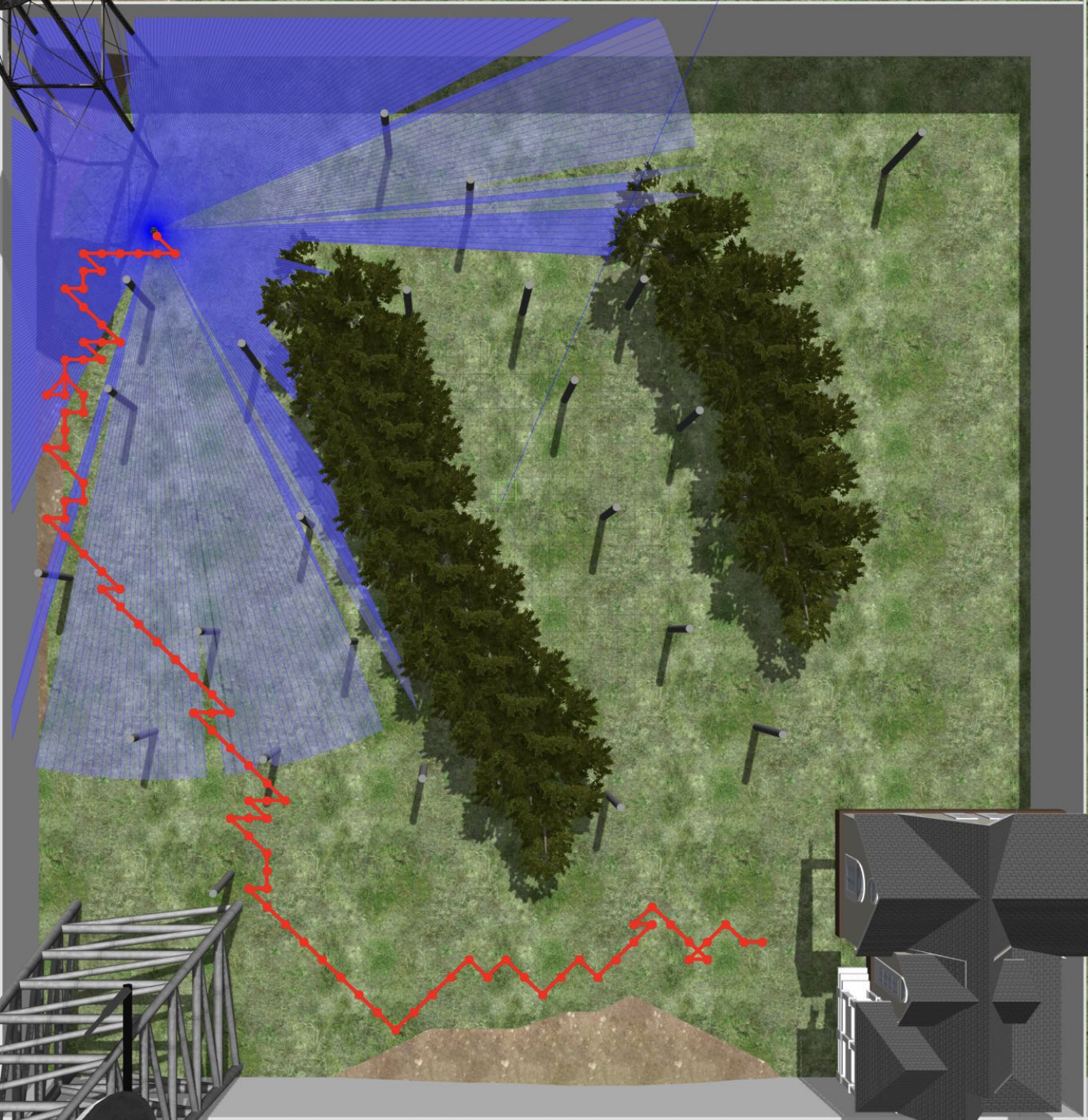}
    \captionsetup{font={small,stretch=1}}
    \caption{
    Trajectory paths (red lines) that UAV traverse in different experiments. 
    Iron bars are randomly distributed in all experiments.
    }
    \label{fig:test_paths}
\end{figure}

As an example, results from six sample experiments are shown in \figref{fig:test_paths}.
The oscillatory patterns observed in the movements of our parsimonious UAV when in 
close proximity to obstacles, as depicted in the figures, distinctly illustrate its 
dynamic obstacle avoidance behavior. 
This behavior becomes particularly evident when the drone's distance from 
obstacles falls below the predefined threshold established during the training 
phase. 

Moreover, we frequently find that certain algorithms excel in specific areas. 
For instance, as demonstrated in \tabref{tab:results}, Dijkstra's algorithm, despite 
achieving a lower success rate, boasts significant time efficiency. 
This variability in performance underscores the importance of allowing users to 
select algorithms tailored to their unique requirements, which highlights the value of 
the modular design of our framework, \algname. 
This design facilitates easy integration and experimentation with emerging RL algorithms, 
offering a versatile platform for future research endeavors.
%
%

\section{Conclusion And Future Works}\label{sec:conclusions}
In this paper, we introduce \textbf{RELAX}, a RL-based autonomous system 
for parsimonious UAVs that carry only one single 2D-LiDAR to successfully perform
navigation in unknown environments.
Rigorous feasibility tests confirm its effectiveness, showcasing a remarkable 
success rate of 90\% in diverse scenarios, outperforming existing algorithms by a
significant margin. 
In addition, we demonstrate \algname's great potential as both RRT and D3QN can be 
replaced by more advanced algorithms and network structures to achieve more 
desirable performance.

Despite the success, ensuring precise 2D-LiDAR detection requires conservative speed 
settings, which limits our system's versatility. 
To mitigate this, we are investigating the integration of multiple 2D-LiDARs 
collecting data from different angles. 
By fusing these data, we aim to counteract the influence of the imperfect 
LiDAR readings, thus further broadening the application potential of \algname.
%
%

\clearpage
\section*{APPENDIX}\label{sec:appendix}

\subsection{Mission Planner Algorithm}\label{alg:mission_planner}
%
%
%
\begin{algorithm}[H]
    \caption{Mission Planner}
    \label{algo:rrt}
    \begin{algorithmic}[1]
    \renewcommand{\algorithmicrequire}{\textbf{Input:}}
    \renewcommand{\algorithmicensure}{\textbf{Output:}}
    \REQUIRE start, target, number of iterations, grid, step size, test range 
    \ENSURE  path between start and target in real environment
    \\ \textit{Initialization} :
    \STATE $N_{start} \gets$ treeNode(start); $N_{target} \gets$ treeNode(end) 
    \STATE $R_{tree} \gets$ RRTAlgorithm($N_{start}$, $N_{end}$, numOfIterations, grid, stepSize, testRange)
    \STATE upperLeftPoint, upperRightPoint, lowerRightPoint, lowerLeftPoint $\gets$ Scanned Occupancy Grid Map; xMinG, xMaxG, yMinG, yMaxG $\gets$ Gazebo World Environment
    \\ \textit{LOOP Process}
    \FOR {$i = 0$ to $numOfIterations$}
        \STATE $R_{tree}$.resetNearestValues(); point $\gets R_{tree}$.sampleAPoint() 
        \STATE $N_{nearest} \gets R_{tree}$.findNearestPoint()
        \STATE $N_{new} \gets R_{tree}$.steerToPoint($N_{nearest}$)
        \STATE flag $\gets$ check if there are obstacles between $N_{new}$ and $N_{nearest}$
        \IF {not reach target}
            \STATE Add $N_{new}$ to $R_{tree}$ and check whether $N_{new}$ in the test range of target, if yes, then break
        \ENDIF
    \ENDFOR
    \STATE $R_{tree}$.WayPoints $\gets$ $R_{tree}$.retraceRRTPath()
    \STATE WaypointsTransformed $\gets$ \textcolor{blue}{Eq.\ref{eq:xnew}, Eq.\ref{eq:ynew}}
    \RETURN WaypointsTransformed 
    \end{algorithmic} 
\end{algorithm}
\vspace{-0.1in}

\subsection{Real-time 2D-LiDAR Filtering Algorithm}
%
%
\begin{algorithm}[H]
    \caption{Real-time 2D-LiDAR Data Filtering}
    \label{algo:lidar_detection}
    \begin{algorithmic}[1]
    \renewcommand{\algorithmicrequire}{\textbf{Input:}}
    \renewcommand{\algorithmicensure}{\textbf{Output:}}
    \REQUIRE LiDAR data from last episode (lidar\_data\_t) and this episode (lidar\_data), a list used for recording function (index\_list)
    \ENSURE  state
    \FOR {$i = 0$ to len(lidar\_data)-1}       
        \IF {lidar\_data\_t[i] - lidar\_data[i] $\geq$ 1.5}
            \STATE
            index\_list[i] $+= 1$
            \STATE $d_{r}$ = floor($0.5*$det\_range)
            \IF {index\_list[i] $\geq d_{r}$}
                \STATE lidar\_data[i] $=$ det\_range $- d_{r} + 1$
                \STATE index\_list[i] $-=$ (det\_range $- d_{r} - 1$)
            \ELSE
                \STATE lidar\_data[i] $=$ lidar\_data\_t[i] $- 1$
            \ENDIF
        \ELSE
            \IF {index\_list[i] $> 0$}
                \STATE index\_list[i] $-= 1$
            \ENDIF
        \ENDIF
    \ENDFOR
    \STATE state = lidar\_data
    \STATE return state
    \end{algorithmic} 
\end{algorithm}
\subsection{Reset Algorithm}
%
%
\begin{algorithm}[htbp]
    \caption{Reset}
    \label{algo:reset}
    \begin{algorithmic}[1]
    \renewcommand{\algorithmicrequire}{\textbf{Input:}}
    \renewcommand{\algorithmicensure}{\textbf{Output:}}
    \REQUIRE a$_{thr}$, b$_{thr}$, offset$_a$, offset$_b$
    \ENSURE  None
    \WHILE{$True$}
        \IF {distance between ($x_c, y_c, z_c$) and ($0,0,4.4$) $\leq 1$}
            \STATE break
        \ELSE
            \IF {($x_t \neq x_c \, \OR \, y_t \neq y_c$)}
                \IF {$\lvert x_t \rvert \geq $a$_{thr}$}
                    \IF {$\lvert x_t \rvert \geq $b$_{thr}$}
                        \STATE $x_t \gets \lvert x_t \rvert -$ offset$_b$
                    \ELSE
                        \STATE $x_t \gets \lvert x_t \rvert - $offset$_a$
                    \ENDIF
                \ELSE
                    \STATE $x_t \gets x_c$
                \ENDIF
                \IF {$\lvert y_t \rvert \geq $a$_{thr}$}
                    \IF {$\lvert y_t \rvert \geq $b$_{thr}$}
                        \STATE $y_t \gets \lvert y_t \rvert -$offset$_b$
                    \ELSE
                        \STATE $y_t \gets \lvert y_t \rvert - $offset$_a$
                    \ENDIF
                \ELSE
                    \STATE $y_t \gets y_c$
                \ENDIF
                \\ \STATE start $\gets [x_t, y_t, 4.4]$
                \STATE Let drone move to start
            \ENDIF
        \ENDIF
    \ENDWHILE
    \end{algorithmic} 
\end{algorithm}
\subsection{Parameters used in Training of RL Agent}
As Deep RL algorithms are sensitive to hyper parameters, we provide the hyper parameters 
we used through our training in \tabref{tab:train_params}. 
%
%
\begin{table}[H]
  \sisetup{group-minimum-digits = 4}
  \centering
  \captionsetup{font={small,stretch=1},justification=raggedright}
  \caption{Parameters used in training of obstacle handler}
  \label{tab:train_params}
  \begin{tabular}{lllS[table-format=5]ll} 
    \toprule
    \thead{Parameter} & \thead{Value} \\
    \midrule
    State dimensions $N_{dim}$ & 11 \\
    Action dimensions $A_{dim}$ & 8 \\
    Training episodes $N_{eps}$ & 500 \\
    Maximum step for one episode 
    $N_{step}$ & 50    \\
    Memory pool size $M$ & $1\times10^{6}$ \\
    Batch size $B$ & 96 \\
    Target network parameter update frequency $f_u$ & 1000 \\
    Discount factor $\gamma$ & $0.99$ \\
    Learning rate $\alpha_l$ & $5\times10^{-4}$    \\
    $\epsilon$-greedy possibility $\max$ $\epsilon_{max}$ & 1.0   \\
    $\epsilon$-greedy possibility $\min$ $\epsilon_{min}$& 0.01   \\
    $\epsilon$-greedy decay factor $\epsilon_{decay}$& $1\times10^{-4}$ \\
    \bottomrule
  \end{tabular}
\end{table}
\bibliography{references}
\bibliographystyle{IEEEtran}

\end{document}